\documentclass[11pt,a4paper]{article}
\usepackage[nohyperref]{emnlp-ijcnlp-2019}
\usepackage{times}
\usepackage{latexsym}

\usepackage{amsmath}
\usepackage{amssymb}

\usepackage{graphicx}
\graphicspath{ {figs/} }
\usepackage{amsmath,amssymb}
\usepackage{eucal}
\usepackage{subcaption}
\usepackage[ruled]{algorithm2e}
\usepackage{url}
\usepackage{multirow}
\usepackage{cleveref}

\aclfinalcopy % Uncomment this line for the final submission

\title{Structured Hierarchical Dialogue Policy with Graph Neural Networks}

\author{Zhi Chen, Xiaoyuan Liu, Lu Chen and Kai Yu \\
  Key Lab. of Shanghai Education Commission for Intelligent Interaction and Cognitive Eng. \\
  SpeechLab, Department of Computer Science and Engineering \\
  Brain Science and Technology Research Center \\
  Shanghai Jiao Tong University, Shanghai, China \\
  {\tt \{zhenchi713,lxy9843,chenlusz,kai.yu\}@sjtu.edu.cn} }

\date{}

\begin{document}
\maketitle
\begin{abstract}
Dialogue policy training for composite tasks, such as restaurant reservation in multiple places, is a practically important and challenging problem. Recently, hierarchical deep reinforcement learning (HDRL) methods have achieved good performance in composite tasks. However, in vanilla HDRL, both top-level and low-level policies are all represented by multi-layer perceptrons (MLPs) which take the concatenation of all observations from the environment as the input for predicting actions. Thus, traditional HDRL approach often suffers from low sampling efficiency and poor transferability. In this paper, we address these problems by utilizing the flexibility of graph neural networks (GNNs). A novel ComNet is proposed to model the structure of a hierarchical agent. The performance of ComNet is tested on composited tasks of the PyDial benchmark. Experiments show that ComNet outperforms vanilla HDRL systems with performance close to the upper bound. It not only achieves sample efficiency but also is more robust to noise while maintaining the transferability to other composite tasks.
\end{abstract}

%In this paper, we address these problems by designing a neural network structure that is more suitable for HDRL-based dialogue management.

\section{Introduction}

%When building a dialogue system, complex tasks that require more information exchange are often more challenging. One example is to handle the restaurant reservation consultation in multiple areas during a single conversation. Specifically, this type of task that needs to complete some subtasks (one area at a time) in order to finish the conversation is called the composite task.

Composite tasks are different from multi-domain dialogue tasks. The latter is often mentioned in papers that focusing on transfer learning. In most case, multi-domain dialogue tasks involve only one domain in a single dialogue, and the performance of this one domain model is tested on different domains in order to highlight its transferability. On the contrary, composite dialogue tasks may involve multiple domains in a single dialogue, and the agent must complete all subtasks (accomplish the goals in all domains) in order to get positive feedback.

Consider the process of completing a composite task (e.g., multi-area restaurant reservation). An agent first chooses a subtask (e.g., reserve-Cambridge-restaurant), then make a sequence of decisions to gather related information (e.g., price range, area) until all information required by users are provided and these subtasks are completed, and then choose the next subtask (e.g., reserve-SF-restaurant) to complete. The state-action space will increase with the number of subtasks. Thus, dialogue policy learning for the composite task needs more exploration, and it needs to take more dialogue turn between agent and user to complete a composite task. The sparse reward problem is further magnified. 

Solving composite tasks using the same method as the one solving single domain tasks may hit obstacles. The complexity of the composite task makes it hard for an agent to learn an acceptable strategy. While hierarchical deep reinforcement learning (HDRL) shows its promising power, by introducing the framework of options over Markov Decision Process (MDP), the original task can be decomposed into two parts: deciding which
subtask to solve and how to solve one subtask, thus simplifying the problem.

However, in previous works, multi-layer perceptrons (MLPs) are often used in DQN to estimate the Q-value. MLPs use the concatenation of the flatten dialogue state as its inputs. In this way, it cannot capture the structural information of the semantic slots in that state easily, which results in low sampling efficiency. In our work, we propose ComNet, which makes use of the Graph Neural Network (GNN) to better leverage the graph structure in the observations (e.g., dialogue states) and being coherent with the HDRL method.

Our main contributions are three-fold: 1. We propose a new framework ComNet combining HDRL and GNN to solve the composite tasks while achieving sample efficiency. 2. We test ComNet based on PyDial \cite{ultes2017pydial} benchmark and show that our result over-performed the vanilla HDRL systems and is more robust to noise in the environment. 3. We test the transferability of our framework and prove that under our framework, an efficient and accurate transfer is possible.

\section{Related Work}

Reinforcement learning is a recently mainstream method to optimize statistical dialogue management policy under the partially observable Markov Decision Process (POMDP) \cite{young2013pomdp}. One line of research is on single-domain task-oriented dialogues with flat deep reinforcement learning approaches, such as DQN~\cite{zhao2016towards,li2017end,chang2017affordable,chen2017agent},policy gradient~\cite{williams2016end,williams2017hybrid} and actor critic~\cite{susample,liu2017iterative,peng2018adversarial}. Multi-domain task-oriented dialogue task is another line, where each domain learns a separate dialogue policy~\cite{gavsic2015policy,gavsic2017dialogue}. 

Recently, \citeauthor{peng2017composite} presented a composite dialogue task. Different from the multi-domain dialogue system, the composite dialogue task requires all the individual subtasks have to be accomplished. The composite dialogue task is formulated by options framework \cite{sutton1998intra} and solved using hierarchical reinforcement learning methods~\cite{budzianowski2017sub,peng2017composite,tang2018subgoal}. All these works are built based on the vanilla HDRL, where the policy is represented by multi-layer perceptron (MLP). However, in this paper, we focus on designing a transferable dialogue policy for the composite dialogue task based on Graph Neural Network \cite{scarselli2009graph}.

 GNN is also used in other aspects of reinforcement learning to provide features like transferability or less over-fitting \cite{wang2018nervenet}. In dialogue system building, models like BUDS also utilize the power of graph for dialogue state tracking \cite{thomson2010bayesian}. Previous works also proved that using GNN to learn a structured dialogue policy can improve system performance significantly in a single-domain setting by creating graph nodes corresponding to the semantic slots and optimizing the graph structure \cite{chen2018structured}. However, for the composite dialogue, we need to exploit the particularity of the tasks and change the complete framework.

\section{Hierarchical Reinforcement Learning}
\label{sect:hrl}
%The learning process of $\pi_{\mathbf{b},g}$ requires intrinsic reward signal given by the evaluation module, which indicates how likely the current subtask is fulfilled. It encourages the agent to solve a subtask before moving on to another subtask. 
Before introducing ComNet, we first present a short review of HRL for a composite task-oriented dialogue system. According to the \emph{options} framework, assume that we have a dialogue state set $\mathcal{B}$, a subtask (or an option) set $\mathcal{G}$ and a primitive action set $\mathcal{A}$.

Compared to the traditional Markov decision process (MDP) setting where an agent can only choose a primitive action at each time step, the decision-making process of hierarchical MDP consists of (1) a top-level policy $\pi_{\mathbf{b}}$ that selects subtasks to be completed, (2) a low-level policy $\pi_{\mathbf{b},g}$ that selects primitive actions to fulfill a given subtask. The top-level policy $\pi_{\mathbf{b}}$ takes as input the belief state $\mathbf{b}$ generated by the global state tracker and selects a subtask $g \in \mathcal{G}$. The low-level policy $\pi_{\mathbf{b},g}$ perceives the current state $\mathbf{b}$ and the subtask $g$, and outputs a primitive action $a\in \mathcal{A}$. The low-level policy $\pi_{\mathbf{b},g}$ is shared by all subtasks.

In this paper, we take two Q-function to represent these two level policies, learned by deep Q-learning approach (DQN) and parameterized by $\theta_e$ and $\theta_i$ respectively. Corresponding to two level policies, there are two kinds of reward signal from the environment (the user): \emph{extrinsic} reward $r^{e}$ and \emph{intrinsic} reward $r^{i}$. The extrinsic rewards guide dialogue agent to choose right subtask order. The intrinsic rewards are used to learn an option policy to achieve a given subtask. The combination of the extrinsic reward and intrinsic reward is to help the dialogue agent to accomplish a composite task as fast as possible. Thus, the extrinsic and intrinsic rewards are designed as follows:

\textbf{Intrinsic Reward.} At the end of a subtask, the agent receives a positive intrinsic reward of 1 for a success subtask or 0 for a failure subtask. In order to encourage shorter dialogues, the agent receives a negative intrinsic reward of -0.05 at each turn.

\textbf{Extrinsic Reward.} Let $K$ be the number of subgoals. At the end of a dialogue, the agent receives a positive extrinsic reward of $K$ for a success dialogue or 0 for a failure dialogue. In order to encourage shorter dialogues, the agent receives a negative extrinsic reward of -0.05 at each turn.

\begin{figure*}[htbp!]
\centering
\includegraphics[width=0.9\textwidth]{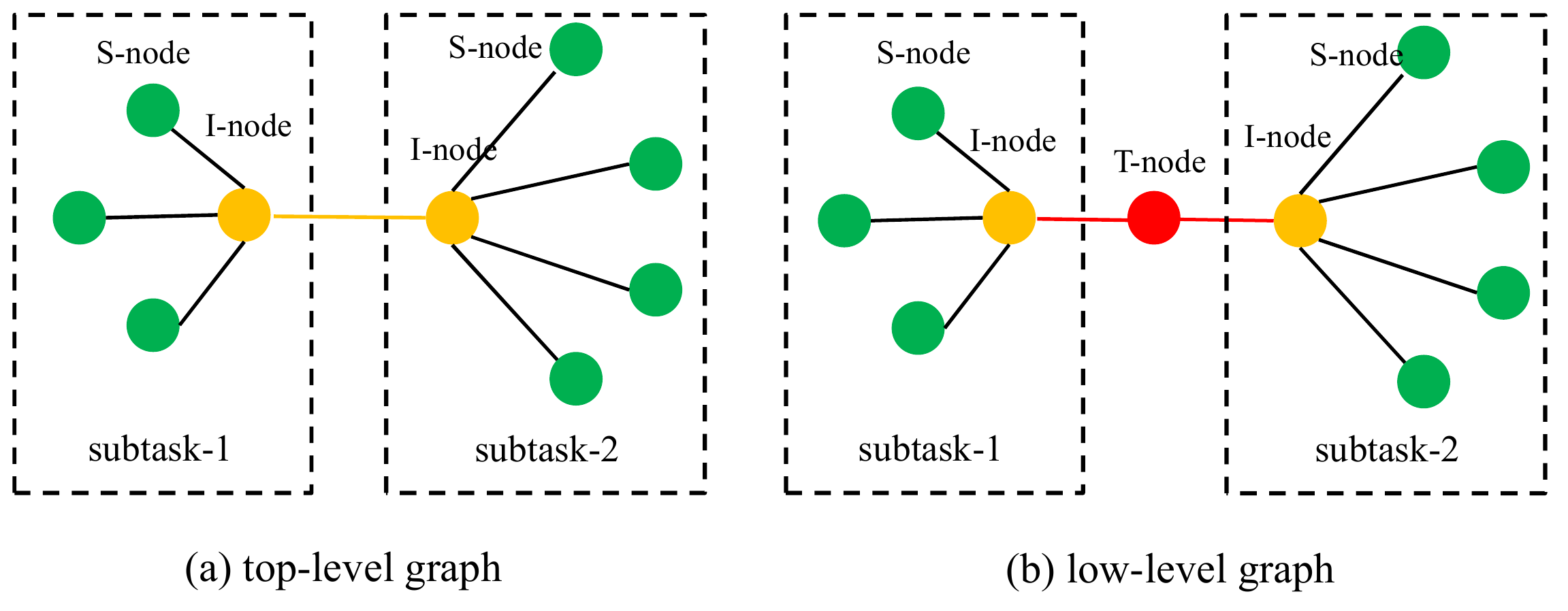}
\caption{A composite dialogue task contains two subtasks, where (a) is the graph of the top-level policy and (b) is the graph of the low-level policy.}
\label{figs:graph}
\end{figure*} 

Assume we have a subtask trajectory of $T$ turns:
$\mathcal{T}_k = (\mathbf{b}_0^k, a_0^k, r_0^k,\dots , \mathbf{b}_T^k, a_T^k, r_T^k)$, where $k$ represents the k-th subtask $g_k$. The dialogue trajectory consists of a sequence of subtask trajectories $\mathcal{T}_0, \mathcal{T}_1. \dots$. According to Q-learning algorithm, the parameter $\theta_e$ of the top-level Q-function is updated as follows:
\begin{small}
\begin{align*}
\theta_e \gets \theta_e + \alpha \cdot (q_k-Q(g_k|\mathbf{b}_0^k;\theta_e))\cdot \nabla_{\theta_e}Q(g_k|\mathbf{b}_0^k;\theta_e), 
\end{align*}
\end{small}
where
\begin{align*}
q_k=\sum_{t=0}^{T}\gamma^{t}r_t^e + \gamma^{T}\max_{g^{\prime}\in \mathcal{G}}Q(g^{\prime}|\mathbf{b}_T^k;\theta_e), 
\end{align*}
and $\alpha$ is the step-size parameter, $\gamma \in [0,1]$ is a discount rate. The first term of the above expression of q equals to the total discounted reward during fulfilling subtask $g_k$, and the second estimates the maximum total discounted value after $g_k$ is completed. 

The learning process of the low-level policy is in a similar way, except that intrinsic rewards are used. For each time step $t=0,1,\dots, T$, 
\begin{small}
\begin{align*}
\theta_i \gets \theta_i + \alpha \cdot (q_t -Q(a_t|\mathbf{b}_t^k,g_k;\theta_i))\cdot \nabla_{\theta_i}Q(a_{t}|\mathbf{b}_{t}^k,g_k;\theta_i), 
\end{align*}
\end{small}
where
\begin{align*}
q_t= r_t^i + \gamma \max_{a^{\prime}\in \mathcal{A}}Q(a^{\prime}|\mathbf{b}_{t+1}^k,g_k;\theta_i). 
\end{align*}

In vanilla HDRL, the above two Q-functions are approximated using MLP. The structure of the dialogue state is ignored in this setting. Thus the task of the MLP policies is to discover the latent relationships between observations. This leads to longer convergence time, requiring more exploration trials. In the next section, we will explain how to construct a graph to represent the relationships in a dialogue observation.  

\section{ComNet}
\label{sect:ComNet}
In this section, we first introduce the notation of the composite task. We then explain how to construct two graphs for two-level policies of a hierarchical dialogue agent, followed by the description of the ComNet. 

\subsection{Composite Dialogue}
\label{sect:comdial}
Task-oriented dialogue systems are typically defined by a structured \emph{ontology}. The \emph{ontology} consists of some properties (or slots) that a user might use to frame a query when fulfilling the task. As for composite dialogue state which contains $K$ subtasks, each subtask corresponds to several slots. For simplification, we take the subtask $k$ as an example to introduce the belief state. There are two boolean attributes for each slot of the subtask $k$, whether it is \emph{requestable} and \emph{informable}. The user can request the value of the requestable slots and can provide specific value as a search constraint for the informable slots. At each dialogue turn, the dialogue state tracker updates a belief state for each informable slot.

Generally, the belief state consists of all the distributions of candidate slot values. The value with the highest belief for each informable slot is selected as a constraint to search the database. The information of the matched entities in the database is added to the final dialogue state. The dialogue state $\mathbf{b}^k$ of the subtask $k$ is decomposed into several \emph{slot-dependent} states and a \emph{slot-independent}
state, represented as $\mathbf{b}^k = \mathbf{b}^{k,1}\oplus \mathbf{b}^{k,2}\oplus \dots \oplus \mathbf{b}^{k,n} \oplus \mathbf{b}^{k,0}$. $\mathbf{b}^{k,j} (1 \le j \le n)$ is the $j$-th informable slot-related state of the subtask $k$, and $\mathbf{b}^{k,0}$ represents the slot-independent state of the subtask $k$. The whole belief state is the concatenation of all the subtask-related state $\mathbf{b}^k$, i.e. $\mathbf{b} = \mathbf{b}^1\oplus \dots \oplus \mathbf{b}^{K}$, which is the input of the top-level dialogue policy.

The output of the top-level policy is a subtask $g \in \mathcal{G}$. In this paper, we use a one-hot vector to represent one specific subtask. Furthermore, the whole belief state $\mathbf{b}$ and the subtask vector $g$ are fed into the low-level policy. The output of the low-level policy is a primitive dialogue action. Similarly, for each subtask $k$, the dialogue action set $\mathcal{A}^k$ can be divided into $n$ slot-related action sets $\mathcal{A}^{k,j} (1 \le j \le n)$, e.g. ${request\_slot}^{k,j}, {inform\_slot}^{k,j}, {select\_slot}^{k,j}$ and a one slot-independent action set $\mathcal{A}^{k,0}$, e.g. ${repeat}^{k,0}, {reqmore}^{k,0}, \dots, {bye}^{k,0}$. The whole dialogue action space $\mathcal{A}$ is the union of all the subtask action spaces. 

\begin{figure*}[htbp!]
\centering
\includegraphics[width=\textwidth]{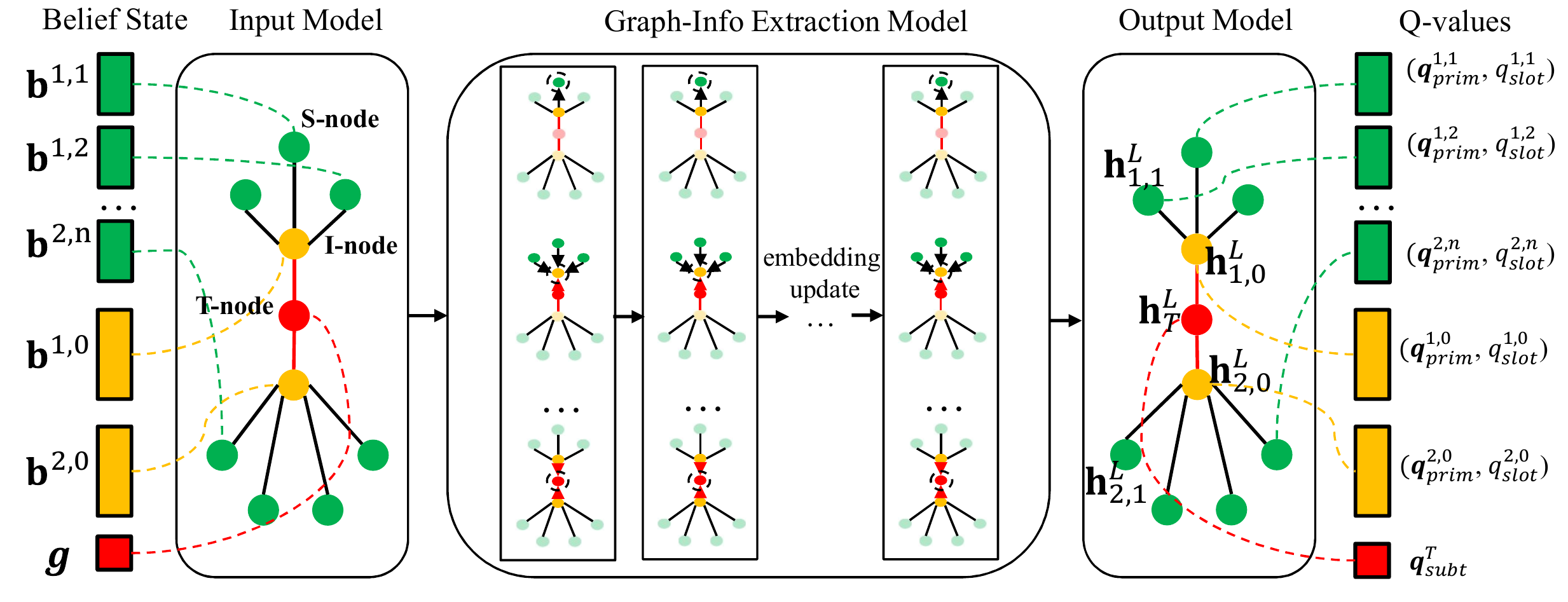}
\caption{The low-level policy model of ComNet contains three parts: input module, graph-info extraction module and output module. In the input module, for each node, ComNet fetches the corresponding elements from the observation. ComNet then computes the messages sent to neighbors in the graph and updates the embedding vector of each node. The main content of the output module is how to use the highest embedding vectors of all the nodes to calculate the corresponding Q-values. The subscript shapes like $(k,i)$ denotes the $i$-th node corresponding to subtask $k$. The top-level policy model of ComNet has a similar structure.}
\label{figs:gnn}
\end{figure*} 

\subsection{Graph Construction}
As introduced in section~\ref{sect:comdial}, the dialogue state $\mathbf{b}$ consists of $K$ subtask-related state, and each subtask-related state can further be decomposed into several slot-dependent states and a slot-independent state which are logically indecomposable, named atomic states. The hierarchical format of the dialogue state can be naturally regarded as a graph. Each node in the graph represents the corresponding atomic state. To simplify the structure of the graph, we choose the slot-independent nodes as the delegate of the nodes which correspond to the same subtask. All the slot-independent nodes are connected with each other in the top-level graph, and the slot-dependent nodes are only connected to their delegate node. Different from the input of top-level policy, the input of low-level policy add a new node named subtask node to represent the goal information, which is produced by the top-level policy. In the low-level graph, the slot-independent nodes are all connected to the subtask node (or the global delegate nodes) instead of connecting to each other.

\subsection{ComNet as Policy Network}
We now turn to ComNet, which parameterizes two-level policies with two Graph Neural Networks (GNNs). Before delving into details, we first introduce our notation. We denote the graph structure as $G=(V,E)$ with nodes $v_i(0\le i\le n)\in V$ and directed edges $e_{ij}\in E$. The adjacency matrix $\mathbf{Z}$ denotes the structure of $G$. If there is a directed edge from $i$-th node $v_i$ to $j$-th node $v_j$, the element $z_{ij}$ of $\mathbf{Z}$ is 1, otherwise $z_{ij}$ is 0. We denote the out-going neighborhood set of node $v_i$ as $\mathcal{N}_{out}(v_i)$. Similarly, $\mathcal{N}_{in}(v_i)$ denotes the in-coming neighborhood set of node $v_i$. Each node $v_i$ has an associated node type $p_i$. Each edge $e_{ij}$ has an edge type $c_e$, which is determined by starting node type $p_i$ and ending node type $p_j$. In other words, two edges have the same type if and only if their starting node type and ending node type are both the same. 

%\texttt{slot-dependent}, \texttt{slot-independent} and \texttt{subtask

For top-level policy, it has two types of nodes: \texttt{slot-dependent} nodes (S-nodes) and \texttt{slot-independent} nodes (I-node). Since there is no edge between  \texttt{slot-dependent} nodes, it has only four edge types. Similarly, for low-level policy, it has three types of nodes (\texttt{slot-dependent}, \texttt{slot-independent} and \texttt{subtask} (T-node)) and four edge types. The two level graphs show in Fig.~\ref{figs:graph}.

Until now, the graphs of top-level policy and low-level policy are both well defined. ComNet, which has two GNNs, is used to parse these graph-format observations of the low-level policy and top-level policy. Each GNN has three parts to extract useful representation from initial graph-format observation: input module, graph-info extraction module and output module.  

\subsubsection{Input Module}
Before each prediction, each node $v_i$ of top-level and low-level graphs will receive the corresponding atomic state $\mathbf{b}$ or subtask information $g$ (represented as $x_i$), which is fed into an input module to obtain a state embedding $\mathbf{h}_i^0$ as follows:
\begin{align*}
\mathbf{h}_i^0 = F_{p_i}(x_i),
\end{align*}
where $F_{p_i}$ is a function for node type $p_i$, which may be a multi-layer perceptron (MLP). Normally, different slots have a different number of candidate values. Therefore, the input dimension of the slot-dependent nodes is different. However, the belief state of each slot is often approximated by the probability of \emph{sorted} top $M$ values ~\cite{gavsic2013gaussian}, where $M$ is usually less than the least value number of all the slots. Thus, the input dimension of nodes with the same type is the same.

\subsubsection{Graph-Info Extraction Module}
The graph-info extraction module takes $\mathbf{h}_i^0$ as the initial embedding for node $v_i$, then further propagates the higher embedding for each node in the graph. The propagation process of node embedding at each extraction layer shows as the following operations.

{\bf Message Computation} At $l$-th step, for every node $v_i$, there is a node embedding $\mathbf{h}_i^{l-1}$. For every out-going node $v_j \in \mathcal{N}_{out}(v_i)$, node $v_i$ computes a message vector as below,
\begin{align*}
\mathbf{m}_{ij}^l = M_{c_e}^l(\mathbf{h}_i^{l-1}),
\end{align*}
where $c_e$ is edge type from node $v_i$ to node $v_j$ and $M_{c_e}^l$ is the message generation function which may be a linear embedding: $M_{c_e}^l(\mathbf{h}_i^{l-1}) = \mathbf{W}_{c_e}^l\mathbf{h}_i^{l-1}$. Note that the subscript $c_e$ indicates that edges of the same edge type share the weight matrix $\mathbf{W}_{c_e}^l$ to be learned. 

{\bf Message Aggregation} After every node finishes computing message, The messages sent from the in-coming neighbors of each node $v_j$ will be aggregated. Specifically, the aggregation process shows as follows:
\begin{align*}
\overline{\mathbf{m}}_{j}^l = A(\{\mathbf{m}_{ij}^l|v_i \in \mathcal{N}_{in}(v_j)\}),
\end{align*}
where $A$ is the aggregation function which may be a summation, average or max-pooling function. $\overline{\mathbf{m}}_{j}^l$ is the aggregated message vector which includes the information sent from all the neighbor nodes.

{\bf Embedding Update} Until now, every node $v_i$ has two kinds of information, the aggregated message vector $\overline{\mathbf{m}}_{i}^l$ and its current embedding vector $\mathbf{h}_i^{l-1}$. The embedding update process shows as below:
\begin{align*}
\mathbf{h}_i^{l} = U_{p_i}^l(\mathbf{h}_i^{l}, \overline{\mathbf{m}}_{i}^l),
\end{align*}
where $U_{p_i}^l$ is the update function for node type $p_i$ at $l$-th extraction layer, which may be a non-linear operation, i.e. 
\begin{align*}
\mathbf{h}_i^{l} = \delta(\lambda^l\mathbf{W}_{p_i}^{l}\mathbf{h}_i^{l} + (1-\lambda^l)\overline{\mathbf{m}}_{i}^l),
\end{align*}
where $\delta$ is an activation function, i.e. RELU, $\lambda^l$ is a weight parameter of the aggregated information which is clipped into $0\backsim 1$, and $\mathbf{W}_{p_i}^{l}$ is a trainable matrix. Note that the subscript $p_i$ indicates that the nodes of the same node type share the same instance of the update function, in our case the parameter $\mathbf{W}_{p_i}^{l}$ is shared.

\subsubsection{Output Module}
After updating node embedding $L$ steps, each node $v_i$ has a final representation $\mathbf{h}_i^{L}$, also represented as $\mathbf{h}_{k,i}^{L}$, where the subscript $k,i$ denotes the node $v_i$ corresponds to the subtask $k$. 

{\bf Top-Level Output}: The top-level policy aims to predict a subtask to be fulfilled. In the top-level graph, for a specific subtask, it corresponds to several S-nodes and one I-node.  Thus, when calculating the Q-value of a specific subtask, all the final embedding of the subtask-related nodes will be used. In particular, for each subtask $k$, we perform the following calculating:
\begin{align*}
q^k_{top} = O_{top}(\sum_{v_i \in S-node}\mathbf{h}_{k,i}^{L}, \mathbf{h}_{k,0}^{L}),
\end{align*}
where $O_{top}$ is the output function which may be a MLP and the subscripts $k,0$ and $k,i$ denote the I-node and $i$-th S-node of the subtask $k$, respectively. In practice, we take the concatenation of $\sum_{v_i \in S-node}\mathbf{h}_{k,i}^{L}$ and $\mathbf{h}_{k,0}^{L}$ as the input of a MLP and outputs a scalar value. For all the subtask, this MLP is shared. When making a decision, all the $q^k_{top}$ will be concatenated, i.e. $\mathbf{q}_{top} = q^1_{top}\oplus \dots \oplus q^K_{top}$, then the subtask is selected according to $\mathbf{q}_{top}$ as done in vanilla DQN.

{\bf Low-Level Output}: The top-level policy aims to predict a primitive dialogue action. As introduced in section~\ref{sect:comdial}, a primitive dialogue action must correspond to a subtask. If we regard slot-independent nodes as a special kind of slot-dependent nodes, a primitive dialogue action can further correspond to a slot node. Thus, the Q-value of each dialogue action contains three parts of information: subtask-level value, slot-level value and primitive value. We use T-node embedding $\mathbf{h}_{T}^{L}$ to compute subtask-level value:
\begin{align*}
\mathbf{q}^T_{subt} = O_{subt}^{T}(\mathbf{h}_{T}^{L}),
\end{align*}
where  $O_{subt}^{T}$ is output function of subtask-level value, which may be a MLP. The output dimension of $O_{subt}^{T}$ is $K$ where each value distributes to a corresponding subtask.
The nodes $v_i$ that belong to S-nodes and I-nodes will compute slot-level value and primitive value:
\begin{align*}
q^{k,i}_{slot} = O_{slot}^{p_i}(\mathbf{h}_{k,i}^{L}), \\
\mathbf{q}^{k,i}_{prim} = O_{prim}^{p_i}(\mathbf{h}_{k,i}^{L}),
\end{align*}
where $O_{slot}^{p_i}$ and $O_{prim}^{p_i}$ are output functions of slot-level value and primitive value respectively, which may be MLPs in practice. Similarly, the subscript $p_i$ indicates that the nodes of the same node type share the same instance of the output functions. The Q-value of an action $\mathbf{a}_{k,i}$ corresponding to the slot node $v_i$ is $\mathbf{q}^{k,i}_{low} = (\mathbf{q}^T_{subt})_k + q^{k,i}_{slot} + \mathbf{q}^{k,i}_{prim}$, where $+$ is element-wise operation and $(\mathbf{q}^T_{subt})_k$ denotes the $k$-th value in $\mathbf{q}^T_{subt}$. When predicting a action, all the $\mathbf{q}^{k,i}_{low}$ will be concatenated, i.e. $\mathbf{q}_{low} = \mathbf{q}^{1,1}_{low}\oplus \dots \oplus \mathbf{q}^{K,0}_{low}$, then the primitive action is chosen according to $\mathbf{q}_{low}$ as done in vanilla DQN.

\begin{table}
\centering
\small
\begin{tabular}{c}
\begin{tabular}{c|lll}
\hline \hline
{\bf Composite Tasks} & {\bf Constraints} & {\bf Requests} & {\bf Values}\\\hline
\verb CR+SFR & 9 & 20 & 904 \\ 
\verb CR+LAP & 14 & 30 & 525 \\
\verb SFR+LAP & 17 & 32 & 893 \\
  \hline \hline
& {\bf Env. 1} & {\bf Env. 2} & {\bf Env. 3} \\ \hline
{\bf SER}  & 0\% & 15\% & 30\% \\
\hline
\end{tabular} \\
\begin{tabular}{l|ccc}
\end{tabular}
\end{tabular}
\caption{The number of data constraints, the number of informative slots that user can request and the number of database result values vary in different composite tasks. Semantic error rate (SER) presents an ascending order in three environments.}\label{tab:env}
\end{table}

\begin{table*}[t!]
\centering
\small
\begin{tabular}{c}
     {\bf User Goal} \\
    \begin{tabular}{lr}
    \begin{tabular}{c}
        {\bf CR}: restaurant\\\\{[}('pricerange', '=', 'dontcare'),\\ ('area', '=', 'centre'),\\ ('food', '=', 'mediterranean'){]} \\\\{\bf req}: {[}'name', 'description'{]}
    \end{tabular}
 &  \begin{tabular}{c}
        {\bf SFR}: restaurant\\\\
        {[}(u'pricerange', '=', u'cheap'), 
        \\(u'allowedforkids', '=', u'dontcare'), 
        \\(u'area', '=', u'civic center'){]}\\\\
        {\bf req}: {[}'name'{]}
         \end{tabular}
      \\ 
    \end{tabular}
\end{tabular}
\begin{tabular}{c|c}
Vanilla HDQN & ComNet \\
\tiny
    \begin{tabular}{p{0.4\linewidth}}
=== CR \\
Sys  : hello(help) \\
User : inform(type="restaurant", pricerange="dontcare", area="centre", food="mediterranean") \\
Sys  : request(area) \\
User : inform(area="centre") \\
Sys  : inform(food="mediterranean", pricerange="expensive", name="la mimosa", area="centre") \\
User : thankyou() \\
Sys  : inform(food="mediterranean", pricerange="expensive", name="la mimosa", area="centre") \\
User : request(description) \\
Sys  : inform(name="la mimosa", description="not available")  \\
=== SFR \\
User : hello(type="restaurant", pricerange="cheap", allowedforkids="dontcare", area="civic center") \\
Sys  : request(area) \\
User : inform(area="civic center") \\
Sys  : request(area) \\
User : inform(area="civic center") \\
Sys  : request(area) \\
User : inform(area="civic center") \\
Sys  : request(area) \\
User : inform(area="civic center") \\
Sys  : request(area) \\
User : bye() \\
Sys  : bye() \\
{\bf *FAILED, subtask-2 is not finished.}\\
    \end{tabular}
    &
    \tiny
    \begin{tabular}{p{0.4\linewidth}}
=== CR \\
Sys  : hello(help) \\
User : inform(type="restaurant") \\
Sys  : request(area) \\
User : inform(area="centre") \\
Sys  : request(food) \\
User : inform(food="mediterranean") \\
Sys  : request(pricerange) \\
User : inform(="dontcare") \\
Sys  : inform(food="mediterranean", pricerange="expensive", name="la mimosa", area="centre") \\
User : thankyou() \\
Sys  : reqmore() \\
User : request(description) \\
Sys  : inform(name="la mimosa", description="not available")  \\
=== SFR \\
User : inform(type="restaurant", area="civic center") \\
Sys  : request(allowedforkids) \\
User : inform(goodformeal="lunch") \\
Sys  : request(allowedforkids) \\
User : inform(allowedforkids="dontcare") \\
Sys  : request(food) \\
User : inform(="dontcare", pricerange="dontcare") \\
Sys  : inform(goodformeal="lunch", name="sai jai thai restaurant", area="civic center", food="thai", allowedforkids="1", pricerange="cheap") \\
User : bye() \\
Sys  : bye() \\
{\bf *SUCCESS} \\
    \end{tabular}
\end{tabular}
\caption{This table is an example of two different trained agents handling the same composite task with same user goal. On the left-hand side, vanilla HDQN succeed in finishing the first subtask but failed to produce meaningful action in the second. On the right-hand side, ComNet successfully full-filled two subtasks, thus succeed in the dialogue.}
\vspace{-0.8cm}
\label{tab:dia}
\end{table*}

\subsubsection{Discussion}
\label{sect:trans}
Note that although the parameters of input module and graph-info extraction module are not shared between the top-level GNN and low-level GNN (shown as Fig.~\ref{figs:gnn}), there are many shared parameters in each single GNN. Assume that now the composite task is changed and one subtask adds some new slot, we only need to create new nodes in each GNN. If the number of the edge type has not changed, the parameters of the GNN will stay the same after adding new nodes. This attribution of ComNet leads to transferability. Generally, if the node type set and edge type set of the composite task $Task_1$ are both subsets of another task $Task_2$'s, the ComNet policy learned in $Task_2$ can be directly used on $Task_1$.

Since the initial output of the same type of nodes has a similar semantic meaning, they share the parameters in ComNet. We hope to use the GNN to propagate the relationships between the nodes in the graph based on the connection of the initial input and the final outputs. 
%If the propagated relationships are task-independent, the ComNet trained on a composite task will be compatible with the other tasks.

\section{Experiments}
In this section, we first verify the effectiveness of ComNet on the composite tasks of the PyDial benchmark. We then investigate the transferability of ComNet.

\subsection{PyDial Benchmark}
A composite dialogue simulation environment is required for the evaluation of our purposed framework. PyDial toolkit \cite{ultes2017pydial} , which supports multi-domain dialogue simulation with error models, has laid a good foundation for our composite task environment building.

We modified the policy management module and user simulation module to support 2-subtask composite dialogue simulation among three available subtasks, which are Cambridge Restaurant (CR), San Francisco Restaurant (SFR) and generic shopping task for laptops (LAP) while preserving fully functional error simulation of different levels in Table \ref{tab:env}. Note that in the policy management module, we discard the domain input provided by dialogue state tracking (DST) module to make a fair comparison. We updated the user simulation module and evaluation management module to support reward design in section~\ref{sect:hrl}.

\subsection{Implementation}
We implement the following three composite task agents to evaluate the performance of our proposed framework.

\begin{itemize}
\item {\bf Vanilla HDQN}: A hierarchical agent using MLPs as its models. This serves as the baseline for our comparison.
\item {\bf ComNet}: Our purposed framework utilizing the flexibility of GNNs. The complete framework is discussed in section~\ref{sect:ComNet}.
\item {\bf Hand-crafted}: A well-designed rule-based agent with a high success rate in composite dialogue without noise. This agent is also used to warm up the training process of the first two agents. Note that this agent uses the precise subtask information provided by DST, which is not fair comparing with the other two.
\end{itemize}

 Here, we train models with 6000 dialogues or iterations. The total number of the training dialogues is broken down into milestones (30 milestones of 200 iterations each). At each milestone, there are 100 dialogues to test the performance of the dialogue policy. 
 \begin{figure}[htbp!]
\centering
\includegraphics[width=0.47\textwidth]{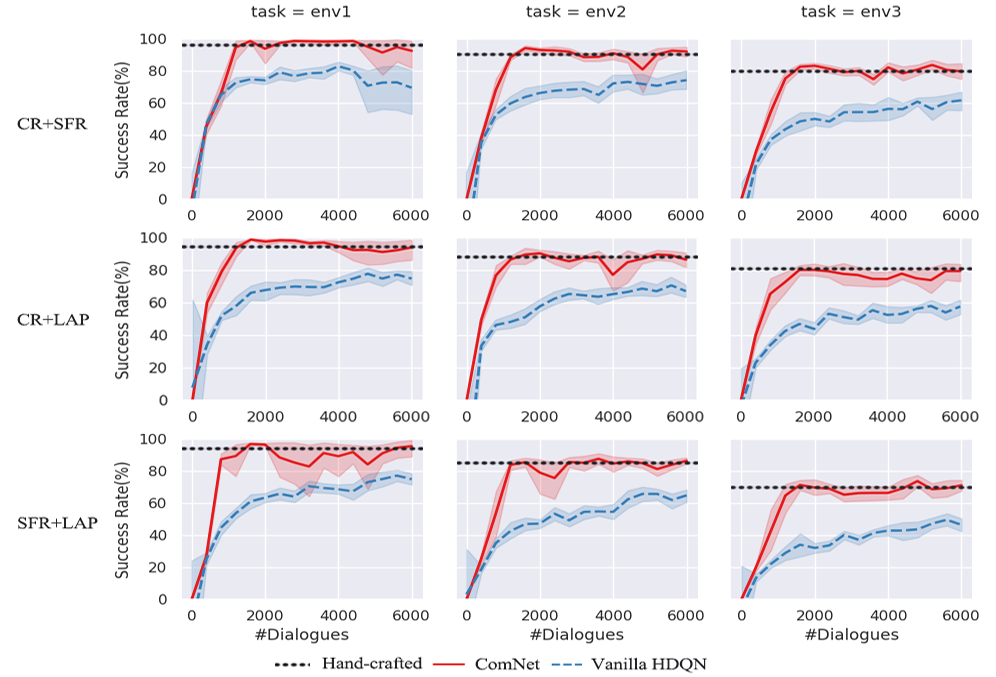}
\caption{The comparison between 3 kinds of agent. ComNet achieved performance close to the upper bound (hand-crafted) while there is still room for improvement for vanilla DQN.}
\label{figs:result1}
\end{figure}
 The results of 3 types of composite tasks in 3 environments in 6,000 training dialogues are shown in Fig.~\ref{figs:result1}.

 \begin{figure*}[htbp!]
\centering
\includegraphics[width=\textwidth]{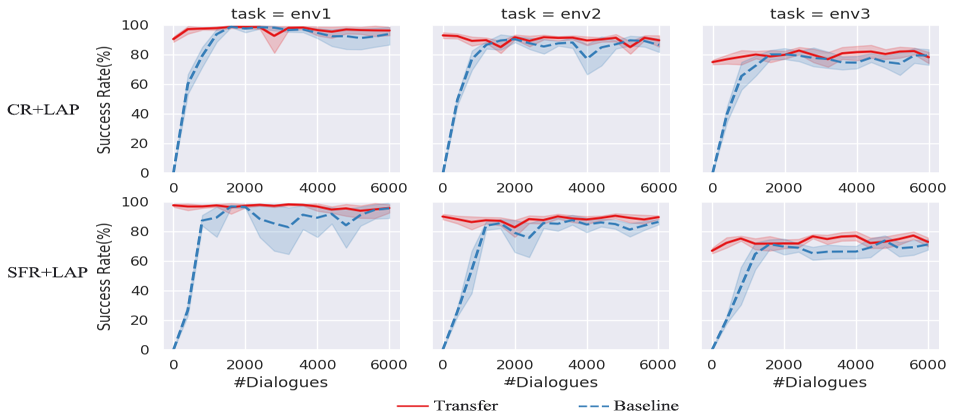}
\caption{The model pretrained on CR+SFR task is compared with the one started with randomized parameters. }
\label{figs:result2}
\end{figure*} 

\subsection{Analysis}
From Fig.~\ref{figs:result1}, we can observe that ComNet outperforms the vanilla MLP policy in all nine settings (3 environments * 3 types of composite tasks) in both success rate and learning speed. In ComNet, the top-level policy and low-level policy are both represented by a GNN where the same of type nodes and the same type of edges share the parameters. It means that the same type of nodes shares the input space (belief state space). Thus the exploration space will greatly decrease. As shown in Fig.~\ref{figs:result1}, ComNet learns to vary faster than vanilla MLP policy.
Note that the hand-crafted agent performs well because it has cheated by peeking at the precise subtask information, which means the hand-crafted agent is solving the multi-domain tasks. This should be the upper bound for the performance of our model. Comparing with vanilla HDQN, our ComNet shows its robustness in all environment by a greater margin, which is helpful for dialogue system building when an ASR or a DST with high accuracy is not available.

We also compared the difference of the dialogues produced by both vanilla HDQN and ComNet after 6000-dialogue training, which is shown in Table \ref{tab:dia}. After that much training, it seems that the vanilla HDQN agent still cannot choose a proper action in some specific dialogue state, which results in the loss of customer patience. On the other hand, ComNet also chose the same action, but it advanced the progress of the dialogue as soon as it got the information it needed, thus finished the task successfully. This also helps to prove that ComNet is more sample efficient comparing to the vanilla framework.

\subsection{Transferability}
As we discussed in section \ref{sect:trans}, another advantage of ComNet is that because of the flexibility of GNNs, ComNet is transferable naturally. 
To evaluate its transferability, we first trained 6,000 dialogues on CR+SFR task. We then initiate the parameters of the policy models on other two composite tasks using trained policy and continue to train and test the models. The result is shown in Fig.~\ref{figs:result2}.

We can find that the transferred model learned on CR+SFR task is compatible with the other two composite tasks. It demonstrates that ComNet can propagate the task-independent relationships among the graph nodes based on the connection of the initial nodes inputs and final outputs. This suggests that it is possible to boost the training process for a new composite task by using pre-trained parameters of related tasks under the framework of ComNet. After all, It is essential to solving the start-cold problems in the task-oriented dialogue systems.

\section{Conclusion}
In this paper, we propose ComNet, which is a structured hierarchical dialogue policy represented by two graph neural networks (GNNs). By replacing MLPs in the traditional HDRL methods, ComNet makes better use of the structural information of dialogue state by separately feeding observations (dialogue state) and the top-level decision into slot-dependent, slot-independent and subtask nodes and exchange message between these nodes. We evaluate our framework on modified PyDial benchmark and show high efficiency, robustness and transferability in all settings.

% \section*{Acknowledgments}

\bibliography{emnlp-ijcnlp-2019}
\bibliographystyle{acl_natbib}

\appendix

\end{document}